# From Rows to Yields: How Foundation Models for Tabular Data Simplify Crop Yield Prediction

Filip Sabo[1], Michele Meroni[1], Maria Piles[2], Martin Claverie[1], Fanie Ferreira[3], Elna Van Den Berg[3], Francesco Collivignarelli[1], Felix Rembold[1]

[1] European Commission, Joint Research Centre (JRC), Ispra, Italy
[2] Universitat de València, Image Processing Laboratory
[3] GeoTerraImage (Pty) Ltd, Pretoria, South Africa

**Abstract**

We present an application of a foundation model for small- to medium-sized tabular data (TabPFN), to sub-national yield forecasting task in South Africa. TabPFN has recently demonstrated superior performance compared to traditional machine learning (ML) models in various regression and classification tasks. We used the dekadal (10-days) time series of Earth Observation (EO; FAPAR and soil moisture) and gridded weather data (air temperature, precipitation and radiation) to forecast the yield of summer crops at the sub-national level. The crop yield data was available for 23 years and for up to 8 provinces. Covariate variables for TabPFN (i.e., EO and weather) were extracted by region and aggregated at a monthly scale. We benchmarked the results of the TabPFN against six ML models and three baseline models. Leave-one-year-out cross-validation experiment setting was used in order to ensure the assessment of the model's capacity to forecast an unseen year. Results showed that TabPFN and ML models exhibit comparable accuracy, outperforming the baselines. Nonetheless, TabPFN demonstrated superior practical utility due to its significantly faster tuning time and reduced requirement for feature engineering. This renders TabPFN a more viable option for real-world operation yield forecasting applications, where efficiency and ease of implementation are paramount.

## 1. Introduction

Crop yield forecasting is a critical component in agricultural planning and management, informing policymakers and stakeholders in the global food supply chain. Early warning and accurate forecasting allow for the anticipation of food production levels, enabling more informed decision-making regarding resource allocation, market strategies, and food security measures[1]. By having access to yield forecasts, stakeholders can better manage risks associated with climate variability, pest outbreaks, and economic fluctuations[2].

Multitemporal earth observation (EO) and gridded weather data are essential for crop yield forecasting over large regions as they inform about the impact of unfavorable weather conditions and reduced biomass development during the growing season that may affect the final crop yield [3,4]. Machine learning (ML) and deep learning (DL) techniques are increasingly applied to process EO data, offering the potential to automate and standardize the yield forecasting process, thereby enhancing accuracy and timeliness [5–12].

Yield forecasting at the sub-national level employs official yield statistics, typically available at some administrative level and for a limited number of years, presenting a significant challenge (i.e. small sample size) for data-driven models. It is even more important in countries with a high risk of food insecurity, such as those in Sub-Saharan Africa. While many of these countries lack existing yield forecasting systems and face data scarcity issues [13], machine learning can still play a significant role. Although machine learning cannot fully overcome the challenge of missing agricultural statistics, it can help rapidly identify optimal methods and make the best use of available predictors in situations where at least a minimum number of years of yield statistics are available.

While DL techniques have achieved remarkable success across various applications, ML algorithms (e.g. Gradient-Boosted Decision Trees) continue to lead in tabular data classification and regression tasks [14]. Although there have been some recent advancements, applying DL techniques to tabular data is still largely considered an "unsolved" challenge.

To provide high accuracy in yield forecasting, ML models (e.g. random forests[15], RF, Gaussian processes[16], GPR, support vector regression[17], SVR, category boosting[18], CatBoost, extreme gradient boosting[19], XgBoost; and Gradient Boosting Regressor[20], GBR, require extensive data curation, including feature selection, feature reduction and feature engineering. Selection and manipulation of raw input data can be performed on the basis of expert and context knowledge (e.g., consecutive dry and wet days from precipitation with different thresholds as in [21,22]) or automatically by testing a large set of predefined options[10]. In both cases, the process is time-consuming and thus challenging to be operationally applied over multiple countries. On the other hand, DL models are more versatile and generally do not require fine-tuning the right input features to excel. In addition, DL models can be used in transfer learning mode for areas where there is not enough data to calibrate the models. Despite these potential attractive features, DL models suffer from the small sample size available for training in the yield forecasting tasks[23]. However, even simple DL architecture models can be demanding in terms of processing and often require training on powerful GPU(s) for hours, which can also be a problem for operational use. This characteristic requires evaluating new approaches or methods, which are less demanding in terms of computing and have a high potential to be used in environments with limited resources.

An interesting development that could address the small data limitation in DL modelling was introduced in [24,25], where a Prior-data Fitted Network (PFN) was introduced to approximate the posterior predictive distribution for any prior from which the data was sampled. The authors selected a transformer-based architecture and trained it on millions of synthetically generated datasets, which provided improved accuracy as compared to various ML benchmarks. Recently, the authors published a second version of their transformer foundation model: the tabular Prior-data Fitted Network (TabPFN)[26]. The authors demonstrated that TabPFN significantly outperformed all the traditional ML (such as SVR, CatBoost, XgBoost, and Light Gradient Boosting Machine) on a small dataset even without extensive tuning. It also performed better than the AutoGluon suite of tuned ML models[27]. This pretrained transformer model is best suited for small to medium tabular datasets, up to 10,000 samples (rows) and 500 features (columns), a size which is ideally suited to sub-national crop statistics data.

Given such promising performance on both the regression and classification tasks of the foundation model for tabular data, for the first time, we evaluated for the time the TabPFN model for a yield forecasting task. We tested the model on summer crops yield forecasting in South Africa, a cereal surplus producing country playing a crucial role in ensuring food security in southern Africa [28].

We benchmarked the TabPFN against three decision tree models, two kernel based models and a linear model. Performances of TabPFN and ML models were also compared to a null model (i.e. simple historical average yield), the trend (linear regression between year and yield) and the PeakFPAR linear model[29] which served as simple baselines. The comparison targeted in-season forecasts in early April, at approximately 75% of the crop summer season. Model performances were assessed in hindcasting while operational forecasts for the year 2024 were compared with the official crop forecasts released monthly by the South African Crop Estimates Committee (CEC).

## 2. Study area and data

We focused on the main summer crops growing in the central and eastern interior summer rainfall region of South Africa: maize, soybeans and sunflowers. All such crops rely largely on rainfed agriculture with only about 8.5% of arable lands irrigated[30], exposing production to climate variability and extremes.

According to the CEC of the Department of Agriculture, Land Reform and Rural Development, maize accounts for 60.8% of the gross value of field crops. In comparison, wheat (grown predominantly in the winter rainfall region of the Western Cape province) contributes 11.5 %, soybeans contribute 5.1% and sunflower contributes 4.5%[31]. The start of summer crop season (SOS) is generally in late October while harvest of Sunflower and soybeans starts in May and Maize ripen naturally on the field and harvested until late July. It is also important to note that South Africa has experienced major droughts in recent years, with major droughts reported in 2015-2016 and in 2018-2020 seasons[30].

The official sub-national yield statistics for South Africa are publicly available from the CEC website (https://www.sagis.org.za/cec_reports.html). The CEC provides forecasts, starting roughly from the end of February. Final yield figures are obtained taking into account also actual deliveries to silos provided by South African grain information service and are published in the next calendar year. These final figures were used as a target label in our modelling. CEC yield forecasts rely on two types of surveys[31]: one conducted via post or e-mail and telephone, and an objective yield survey (in-field measurements) specifically for maize and wheat. For summer crops, a trend analysis is utilized, while crop modeling is exclusively applied to maize. Area estimates are derived from a statistics-based aerial survey along with two additional surveys conducted via post, e-mail, and telephone. The final estimates result from a consensus decision among CEC analysts.

The time series of crop yield data used in this study spaned the period 2001 – 2023. Although a longer archive of crop yield records was available, we were limited to 2001 because we use data from the Moderate Resolution Imaging Spectroradiometer (MODIS) instruments, which are available from mid 2000 on. From the original set of provinces we discarded those having a very marginal share of national area for each of the three crops (i.e. less than 0.5 % of total crop specific area). Therefore, data was available for a total of 23 years and covered 5 to 8 regions, depending on the crop. This brought the total number of labelled data to 184 for maize, 138 for soybeans, and 115 for sunflowers.

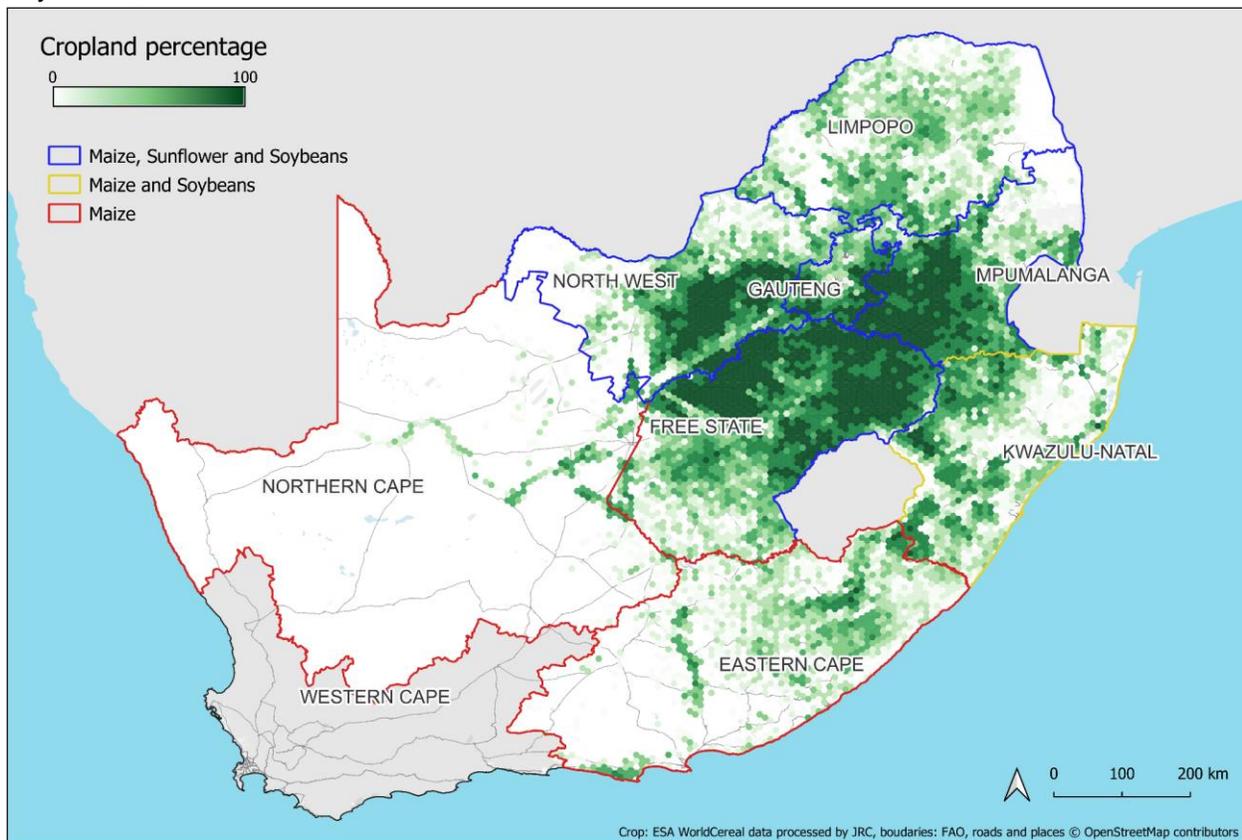

*Figure 1. Overview of the summer cropland distribution in South Africa. Cropland percentage cover sourced from European Space Agency WorldCereal layer. Province boundaries are colored based on the presence of the crops in the yield statistics.*

Environmental data (i.e. EO and weather data) were sourced from the open and public Joint Research Centre (JRC) ASAP Early warning system (https://agricultural-production-hotspots.ec.europa.eu/),[32,33]. We extracted dekadal (10-days) time series of EO data (Fraction of Absorbed Photosynthetically Active Radiation, FPAR; soil moisture), and gridded weather data (temperature, precipitation and radiation) for each province. The data set incorporated FPAR as an indicator of green biomass, quantitative information on soil wetness, and three primary meteorological variables that impact crop growth: precipitation, temperature, and global radiation. Sources and details of the EO and meteorological data used are given in Table 1. Pixel level values of input variables were spatially aggregated to the province level as the weighted average according to fractional area occupied by crops in each 500 m pixel. As crop type specific crop masks are not available for all the three crops, we used the one from the WorldCereal project[34] for maize. Although suboptimal for soybeans and sunflower, it is a pragmatic compromise considering that the three summer crops are grown in the same areas and are typically subjected to crop rotation.

Table 1. Input Earth Observation (EO) and Meteorological (Meteo) data.

| Input type | Variable | Spatial resolution | Time span (from) | Source |
|---|---|---|---|---|
| EO | FPAR | 500m | 2001 | MODIS/VIIRS ([35]) |
|  | Soil moisture | 25km | 1978 | Copernicus Climate Change Service[36] |
| Meteo | Precipitation | 5km | 1984 | CHIRPS v2[37] |
|  | Temperature | 25km | 1984 | ECMWF[32] |
|  | Solar radiation | 25km | 1984 | ECMWF[32] |

The dekadal time series data were aggregated at the monthly time step for each month over the growing season (October to May) by computing statistics such as the average, maximum, minimum, and sum, depending on the variable. In total, 14 sets with a selection of features and summarizing metrics were defined as suitable candidates for the ML pipeline (see Table 2).

Table 2. Monthly aggregation of EO (FPAR, soil moisture) and weather data (solar radiation, Rad; precipitation, Rain; and temperature, T). The minus symbol following the set name indicates a reduced (i.e. less variables) version of the set.

| Set name | Earth observation | | | Meteorology | | | | |
|---|---|---|---|---|---|---|---|---|
|  | FPAR | | Soil moisture | Rad | Rain | T | T | T |
|  | avg | max | avg | sum | sum | avg | min | max |
| RS Met SM | • | • | • | • | • | • | • | • |
| RS Met | • | • |  | • | • | • | • | • |
| RS | • | • |  |  |  |  |  |  |
| Met |  |  |  | • | • | • | • | • |
| maxRS Met |  | • |  | • | • | • | • | • |
| maxRS Met- |  | • |  |  | • | • |  |  |
| maxRS Met- SM |  | • | • |  | • | • |  |  |
| maxRS |  | • |  |  |  |  |  |  |
| RS Met SM- | • |  | • |  | • | • |  |  |
| RS Met- | • |  |  |  | • | • |  |  |
| RS- | • |  |  |  |  |  |  |  |
| RS- SM- | • |  | • |  |  |  |  |  |
| Met- SM |  |  | • | • | • | • |  |  |
| Met- |  |  |  | • | • | • |  |  |

In addition to time series data, we incorporated a province code as a categorical input to enhance the ML models' predictive accuracy[10]. This input helps the model accounting for unobserved factors, e.g. variations in management practices or soil properties across regions. The categorical variables were directly fed into TabPFN, which encodes categorical data natively. In the case of ML models, the categorical data were converted into numerical features using one-hot encoding and then included as additional inputs.

Finally, we also opted for including a yield trend estimation in the predictor set to account for observed time trends in various provinces (all for maize, three for soybeans, none for sunflowers; at 0.05 significance level). Ascertaining and removing the existence of a yield trend (e.g. due to technological improvements or increased availability and use of fertilizers) and modelling it (e.g. using a linear or quadratic time trend) is challenging and typically requires an in-depth analysis of the yield statistics. Here, we chose a pragmatic approach suited for the nested calibration we employ and the intended operational use. Trend estimation for year Y was made using a Theil-Sen linear estimator using the previous 12 years of data.

## 3. Methods

### 3.1 TabPFN

TabPFN is a transformer model trained on millions of synthetically generated datasets and it performs in-context learning (ICL), a characteristic that allows pre-trained models, transformer model in this case, to learn not only simple algorithms but also more complex ones such are Gaussian processes and Bayesian neural networks. Therefore, the ICL makes the TabPFN well-suited for a wide range of classification and regression tasks. Furthermore, TabPFN possesses several advantageous properties that greatly facilitate its deployment for near real-time operational purposes: i) handles missing values, ii) can model uncertainty, iii) provides feature explainability with a built-in library[38], and iv) a single forward pass allows for fast training and inference. The authors also demonstrated that TabPFN is robust to uninformative features, outliers and even to missing categorical features. All of these properties make TabPFN a strong candidate for the operational crop yield forecasting task.

We have evaluated the performances of the TabPFN in two different setups. The first one is the default one, where the model performs a single forward pass for training and inference. The second setup used the post hoc ensemble (PHE), where TabPFN used a tuned ensemble to enhance performance. The configurations for PHE were predefined and, by default, the models were evaluated using a holdout set or a cross validation with early stopping.

### 3.2 Benchmark and baseline models

We evaluated the performance of TabPFN models against a wide range of traditional machine learning models currently used in the ASAP JRC pre-operational yield forecasting pipeline[10,23]. Our comparison included three decision tree models such as XGBoost, GBR, and Random

Forest; kernel-based models like Gaussian Process Regression and SVR with linear and radial basis function (SVR lin and SVR rbf); and the Least Absolute Shrinkage and Selection Operator (LASSO) linear model. In addition, three simple baseline models were included in the comparison: 1) the null model, which uses the average of observed yields per administrative unit; 2) the peak FPAR model, a simple but effective approach[29] which forecasts yield based on the linear regression between maximum FPAR and yield at the administrative unit level; and 3) the trend model, which applies linear regression between year and yield[10].

## 4. Experiments

For the ML models, we started the modelling workflow by evaluating our manually defined feature sets of Table 2. For each feature set, after z-scoring all input features, we tested different pipeline configurations. A configuration is defined by: the ML model used, additional input features passed (trend, one hot encoded (OHE) admin unit id) and additional optional feature selection (Minimum Redundancy and Maximum Relevance, MRMR[39]) and data reduction (Principal Component Analysis, PCA). All possible combinations of configuration options were tested (n=96) while tuning the specific model hyperparameters at the same time. To avoid information leakage, we applied a nested leave-one-year-out (LOYO) cross-validation where the model error metrics were derived from the outer loop (i.e. on the test set) while model hyperparameters setting and best model selection was performed in the inner loop (i.e. on the validation set). It is noted that in this way we selected and assessed the performances of a pipeline rather than a model, i.e. in operation, the identified configuration is tuned on all available data to forecast the current season yield.

In contrast to ML models, TabPFN does not require feature selection, data reduction nor hyperparameter tuning in the default mode. The training and inference are done in a single forward pass. This makes the computational and time requirements very minimal. TabPFN PHE, on the other hand, is used to enhance the performance by automatically ensembling and tuning a predefined number of TabPFN models. The input feature set for TabPFN and TabPFN PHE contained all the features. For the operational forecasts, TabPFN models are simply trained on all the available data and applied on the input feature data from the forecasting year. The simplified general workflow is shown in Figure 2.

To make the best use of the small sample size, we have modified the original PHE cross-validation setup so that the data split was done with LOYO for validation as it was done in the tuning phase of the ML pipeline[10]. Leaving a full year out, or reserving a number of years for the test set is essential to assess the real-world performances of the yield forecasting task where the model has no information about the yield for the season being forecasted.

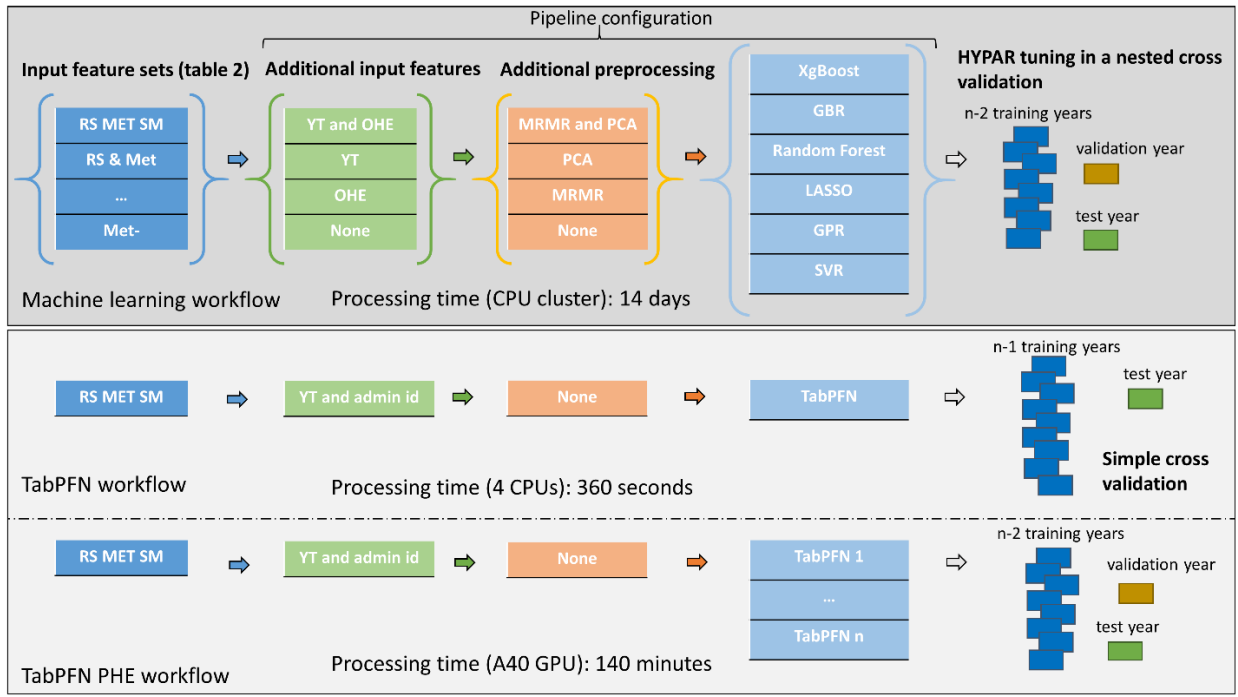

*Figure 2 Simplified machine learning (ML) and TabPFN workflows. YT corresponds to Yield Trend, OHE to one hot encoding of admin code, MRMR is the Minimum Redundancy Maximum Relevance while PCA is the principal component analysis. Nested leave-one-year-out (LOYO) cross validation and hyperaparameter (HYPAR) tuning was used for ML and TabPFN PHE models while a simple LOYO cross validation was used in the case for TabPFN. TabPFN PHE additionally uses model ensembling.*

## 5. Model evaluation

The ML and TabPFN workflows were evaluated through hindcasting for a prediction made at 75% of the crop growing season (in early April) over the period 2002 - 2023. This means that for each left out year, the hindcasting experiments used the data available in time until 75% of the summer crop season. Overall relative root mean square error in prediction was calculated between modelled and observed yield and normalized by the mean yield per crop (rRMSEp). The same input features were used for ML and TabPFN workflows, i.e. the full set of variables described in Table 2 plus administrative unit identifiers and yield trend.

In order to assess the differences between the rRMSEp group means, for all the models and crops, we performed an analysis of variance (ANOVA) followed by Turkey's Honestly Significant Difference (HSD, at 0.05 significance level) test for multiple comparisons. Furthermore, feature importance plots based on Shapley values[38] were also shown for TabPFN model.

In addition, as an example of fully operational application, we forecasted the final yield of the three summer crops at 75% of the growing season (i.e. beginning of April 2024) and compared it with the official CEC forecasts issued from late February throughout the end growing season. The forecasted yield for TabPFN was shown alongside the 95% uncertainty intervals.

## 6. Results and discussion

Hindcasting comparative performances are shown in Figure 3 for the baseline, benchmark and TabPFN models. For all the three crops, both ML, TabPFN and TabPFN PHE models outperformed the simple baseline models. Remarkably, TabPFN, run in a single forward pass for a total of around 360 seconds (utilizing 4 CPUs) per crop, performed only slightly worse than the best ML model configuration for all the crops. TabPFN PHE tuned for 138 mins performed almost the same as the TabPFN and ML for all the crops. Allowing tuning runs longer than 138 mins was not beneficial. Indeed, with the early stop setting in place the PHE process often finished before the allotted time was exhausted. For maize, the TabPFN had an rRMSEp of 8.9%, TabPFN PHE had an rRMSEp of 9.37% and the ML approach had an rRMSEp of 7.39%. For soybeans, the rRMSEp for TabPFN, TabPFN PHE and ML models was 15.1%, 14.81% and 13.51%, respectively. In the case of sunflower, the most accurate ML model achieved an rRMSEp of 13.59%, while both the TabPFN models performed only slightly worse with an rRMSEp of 15.04% and 14.93%.

ANOVA analysis has shown that, for maize, there were no significant differences between the ML, TabPFN, and TabPFN PHE average rRMSEp, all denoted with the letter 'a'. These models were significantly different from Null (denoted as group 'b') and PeakFPAR (group 'c'). The Trend group overlapped with both groups 'a' and 'c', indicating no significant difference with both. This latter result is likely because a significant linear trend was observed in the data for this crop, making the trend model a competitive estimator for maize. For soybeans, ML, TabPFN, and TabPFN PHE (group 'a') were not significantly different from each other. Benchmark models were all significantly different from the group 'a'. For sunflowers, no significant differences were detected among the models.

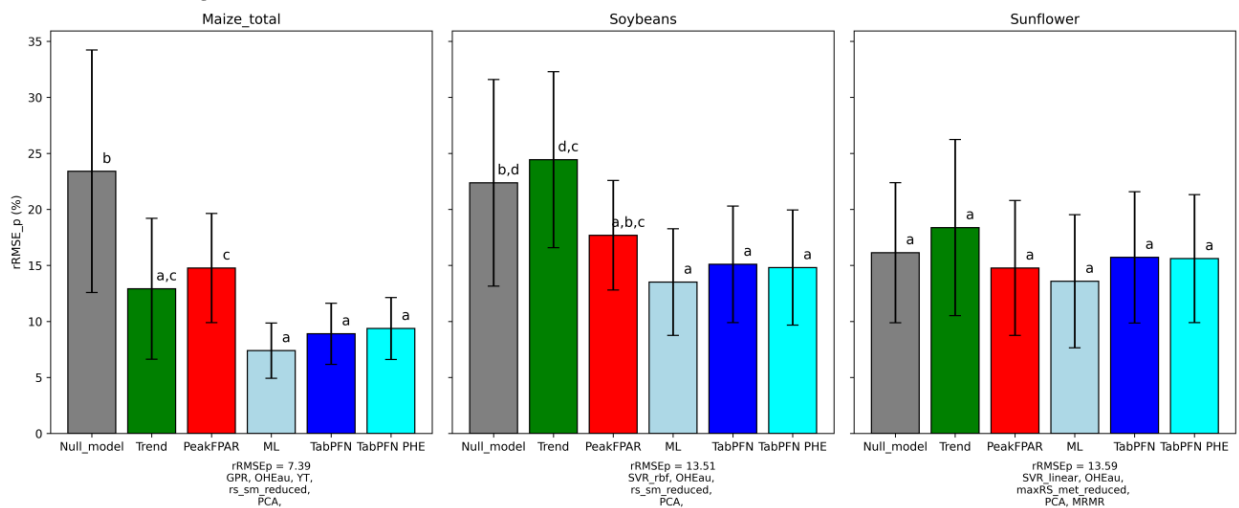

*Figure 3. Comparative performance (average rRMSE and standard deviation) of the benchmark, machine learning (ML) and TabPFN models. The best ML model is labelled with the selected configuration pipeline from the hindcasting (figure 2). Significant differences among means are denoted by different letters (Tukey's HSD test, p = 0.05). Models with the same letter are not significantly different from each other. For Sunflowers, the ANOVA test is not significant (no significant differences among the groups).*

Reasons for this could be related the fact that sunflowers occupy only a minor part of the cropped area (13% against 23 and 64% of soybeans and maize, respectively) and thus have less representative EO and meteorological features due to the use of a single crop mask specific to maize. Additionally, sunflower, being a moderately drought tolerant crop, shows the lowest interannual variability in the yield statistics, likely making its yield more difficult to be modelled with our environmental variables.

It is interesting to note that the TabPFN PHE approach did not improve the accuracy of the predictions.

Finally, popular decision-tree models (XgBoost, GBR and Random Forest) never outperformed the kernel based GPR and SVR models in any of the scenarios. SVR and GPR have consistently outperformed the decision tree based models in line with previous studies in different geographical settings[10,23] where the best performing model was found to be SVR.

Feature importance plots based on Shapley values are shown for TabPFN model in figure 4. For maize, the most important feature was the yield trend followed by the maximum FPAR value from month 5 (March) and categorical variable admin_name, confirming the presence of a yield trend while indicating the importance of biomass development and of among-regions differences. For soybeans, the most important features are all similar as for maize (admin_name, yield trend and FPAR) with the exception incident radiation for month 4 (February), ranking second for soybean. The importance of incident radiation feature on the final yield can be tentatively explained by the presence of more cloudy days (resulting in lower radiation) and potentially increased rainfall.

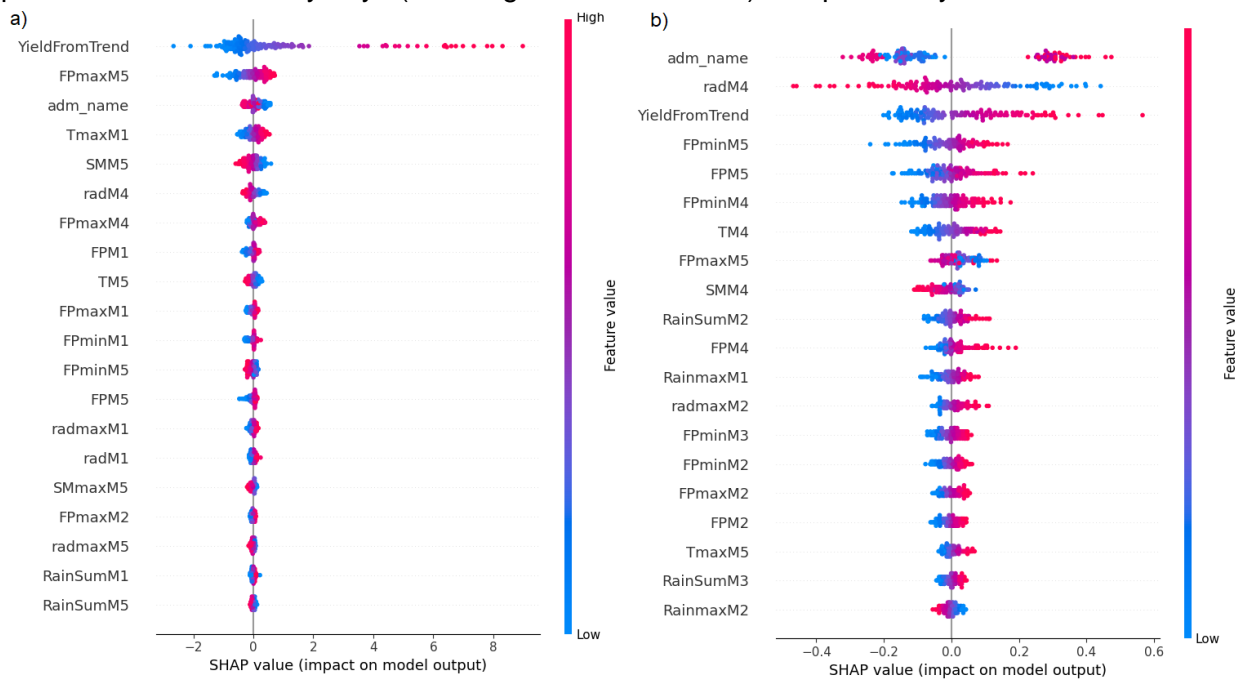

Figure 4. SHAP values sorted by the most important features for a) maize and b) soybeans. Feature names are based on table 2 followed by the month (M). Adm_name is the categorical variable region ID.

After evaluating the workflow in hindcasting, we further evaluated the models in fully operational settings during the 2024 summer crop season, forecasting the final yield at 75% of the growing season (early April) for the two crops where the ML and TabPFN significantly outperformed the benchmark models, i.e. maize and soybeans. The best selected ML model, in terms of RMSE in validation (see Figure 3 for model specifications), was used for operational crop yield forecasts at subnational level. TabPFN was run in its default setup with a single forward pass. The operational forecasts were compared with official CEC forecasts issued at the end of February, then nearly at the same time of the forecasts (end of March for CEC vs early April for ML and TabPFN) and the updates issued later on up to well beyond the end of the season (August). Operational forecasts for maize and soybeans for 2024 are shown in figures 5 and 6 together with TabPFN 95% uncertainty intervals.

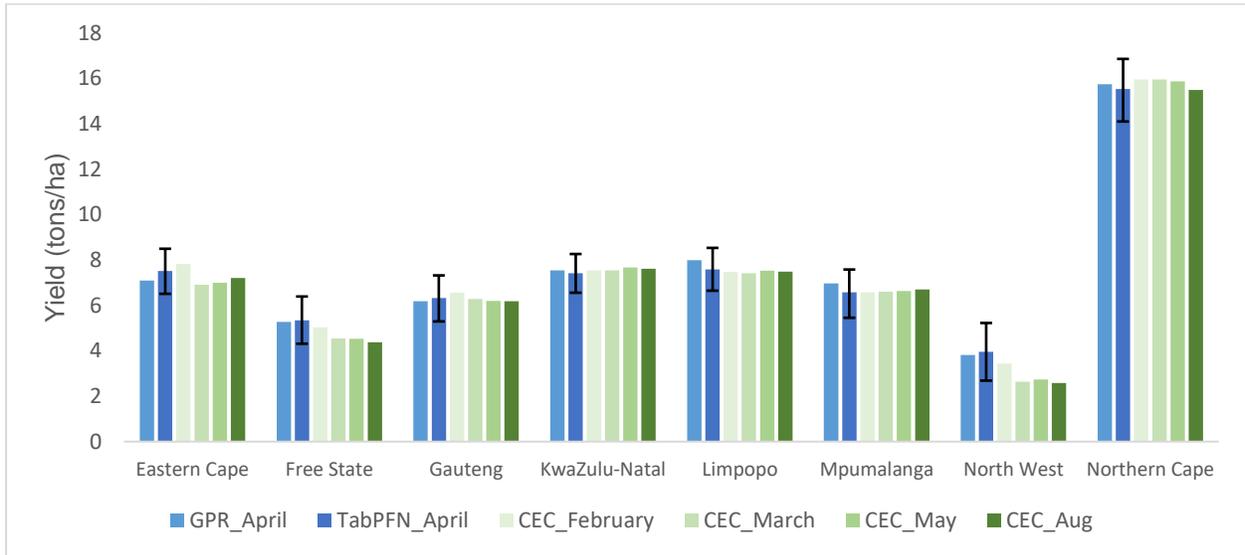

*Figure 5. Gaussian Process Regression (GPR) and TabPFN early April 2024 maize yield forecasts over the main producing provinces compared to CEC estimates starting end of February up to August. Error bars represent TabPFN 95% confidence intervals.*

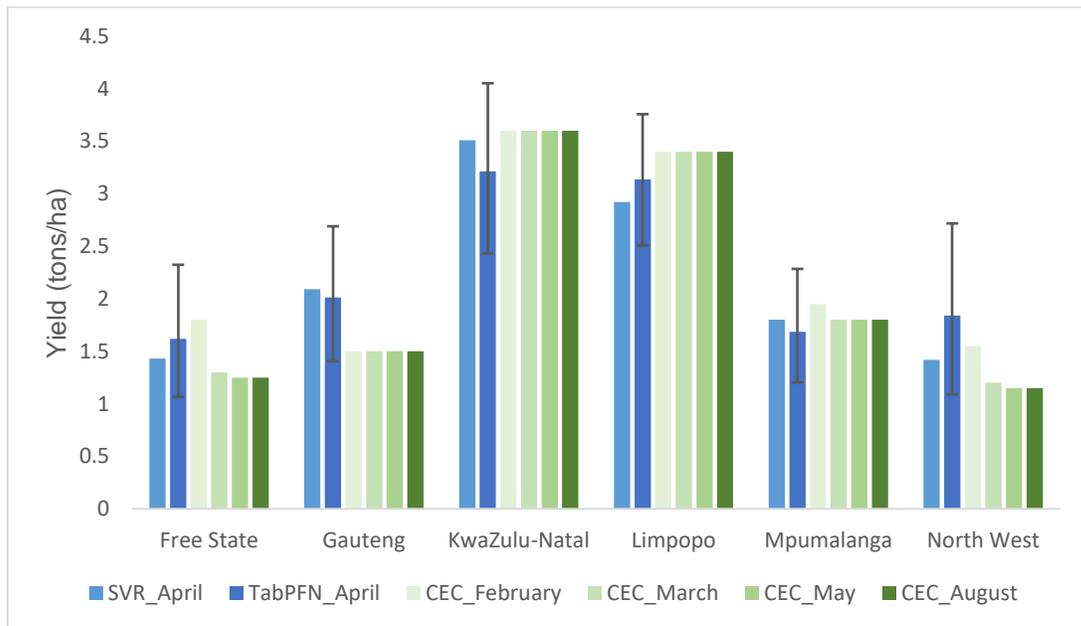

*Figure 6. ML and TabPFN early April 2024 soybeans yield forecasts over the main producing provinces compared to CEC estimates starting end of February up to August. Error bars represent TabPFN 95% confidence intervals.*

Both the ML and TabPFN captured the spatial variability of yields across South African provinces, going from low yield in the North West to very high yields in the Northern Cape. Both ML and TabPFN provided similar predictions for the 2024 season and were aligned with the CEC March predictions with average percentage difference of 10% for maize and 22% for soybeans. The largest difference between TabPFN and CEC estimates was found in the North West (NW) province, both for maize and soybeans, with forecasted yield larger than the CEC estimate with around 50% percentage difference for both crops. The uncertainty analysis for TabPFN showed that the yield values from the CEC for the NW province were not within the 95% confidence interval in the case of maize. In the case of soybeans, all the CEC yield values fell within the 95% confidence interval, however, the percentage of mean uncertainty was higher compared to maize. It is interesting to observe that the CEC yield was below average only in NW province while environmental variables, including FPAR were close to average. In agreement with hindcasting results (figure 3), forecasts for soybeans had larger differences compared with CEC, especially for Gauteng and NW provinces.

Forecasts in 2024 exemplify the distinct advantages of the automated procedures for yield estimation using ML and DL models. These methods can leverage publicly available remote sensing data, such as those provided by ASAP and can be implemented during the season and at large scale with reasonable accuracy. In countries without an effective yield forecasting system such as the one of CEC, these modelling approaches may provide the only prospective source. For the case of South Africa, ML and DL forecasts feed the CEC crop yield estimation process with third-party independent information.

We note that the ML models went through a time consuming feature selection, feature reduction and hyper parameter optimization in order to select the best model. For the entire ML hindcasting procedure to finish it took 14 days running on a JRC HtCondor CPU cluster[40] (having access to about 500 nodes on average) while the TabPFN PHE training and prediction only took approximately 2 hours running on a server equipped with one A40 GPU. TabPFN in default mode, having access to 4 CPUs, took only about 360 seconds (20 seconds when using one A40 GPU) in total for training and prediction.

Considering performance versus resource requirements (cost), the TabPFN is overwhelmingly the preferred option given that, not only does it not require any time-consuming feature engineering and hyperparameter optimization, but its accuracy is also equivalent for in-season forecasts and better for end of the season forecasts compared with ML (data not shown). Low resource demand and simplicity in setting up the processing chain, along with free availability of environmental predictors (e.g. from the ASAP system) may help increase the take up from non-specialised governmental institutions in low to middle income countries. In addition, TabPFN is more fit for the purpose of an operational yield forecasting system that operates on multiple countries and requires recalibration each year when new yield statistics are made available.

In addition, its built-in modules —such as explainability, ability to model uncertainty, and robust handling of missing values and outliers—make it a strong candidate to custom-designed machine learning pipelines, particularly in tasks involving small tabular datasets. In particular, in an operational context, explainability and uncertainty estimates can play a key role in improving users' understanding of the model outcomes and thus their utilization in decision making.

## 7. Conclusion

For the first time, we have evaluated the performances of a pre-trained tabular foundation model (TabPFN) for forecasting crop yield at the sub-national level. We focused on maize, soybeans and sunflower crops in South Africa, a major maize producer in Southern Africa, often referred to as the region's "grain basket". South Africa is also a major exporter of maize to neighboring countries, making it a key player in the region's food security. We compared the historical performance of TabPFN against various machine learning models and baseline non-ML approaches and tested the workflow in a fully operational setting for the 2024 summer crop season.

Results showed that TabPFN and ML models exhibit comparable accuracy, outperforming the baselines (statistically significant for maize and soybeans). Nonetheless, TabPFN demonstrates superior practical utility due to its significantly faster tuning time and elimination of feature engineering requirements. This renders TabPFN a more viable option for real-world operational yield forecasting applications, where efficiency and ease of implementation are paramount. An additional benefit of TabPFN is its ability to process raw data without requiring extensive testing of input model sets, optional inputs, feature selection, and data reduction. This simplifies the coding process, making it more accessible and user-friendly, which is expected to lower the barriers for national stakeholders to adopt and integrate the modeling system into forecasting workflows.

Surprisingly, the TabPFN PHE option, which uses several forward passes and model ensembling, did not perform better than the default TabPFN, which operates with a single forward pass and without ensembling.

Building on the present study, our future work will expand the scope of this comparison to several African countries, where we are developing an operational sub-national yield forecasting system. The objective of this forthcoming research is to confirm the scalability and sustainability of the TabPFN approach, paving the way for its potential adoption as a standard modeling approach for yield forecasting in these countries.

## Code availability

Python code for machine learning models are available on the JRC GitHub repository:
https://github.com/ec-jrc/ml4cast-ml
TabPFN code is also available on GitHub:
https://github.com/PriorLabs/TabPFN